\documentclass{sig-alternate_arXiv}

\usepackage{times}
\usepackage[numbers]{natbib}
\usepackage{color}
\usepackage{epsfig}
\usepackage{amsmath}
\usepackage{amssymb}
\usepackage{wrapfig}
\usepackage{float}
\usepackage{algorithm}
\usepackage{algorithmic}
\usepackage{enumitem}
\usepackage{defs_icml15}
\usepackage{graphicx} % more modern
\usepackage{subfigure} 
\usepackage{fancyhdr}

\newcommand{\sml}[1]{{\scalebox{0.6}{{({$#1$})}}}}

\begin{document}
%
% --- Author Metadata here ---
%\conferenceinfo{KDD}{'15 Sydney, Austrailia}
%\CopyrightYear{2007} % Allows default copyright year (20XX) to be over-ridden - IF NEED BE.
%\crdata{0-12345-67-8/90/01}  % Allows default copyright data (0-89791-88-6/97/05) to be over-ridden - IF NEED BE.
% --- End of Author Metadata ---

\title{Large-Scale Distributed Bayesian Matrix Factorization\\
using Stochastic Gradient MCMC}

%\titlenote{(Produces the permission block, and
%copyright information). For use with
%SIG-ALTERNATE.CLS. Supported by ACM.}}
%\subtitle{[Extended Abstract]
%\titlenote{A full version of this paper is available as
%\textit{Author's Guide to Preparing ACM SIG Proceedings Using
%\LaTeX$2_\epsilon$\ and BibTeX} at

\numberofauthors{5} %  in this sample file, there are a *total*
\author{
% 1st. author
\alignauthor
Sungjin Ahn\\
%\titlenote{Dr.~Trovato insisted his name be first.}\\
%       \affaddr{Dept. of Computer Science}\\
       \affaddr{University of California, Irvine}\\
       \email{sungjia@ics.uci.edu}
% 2nd. author
\alignauthor
Anoop Korattikara
\titlenote{A. Korattikara contributed to this work while he was at the UC Irvine.}\\
       \affaddr{Google}\\
       \email{kbanoop@google.com}
% 3rd. author
\alignauthor Nathan Liu\\
%\titlenote{This author is the
%one who did all the really hard work.}\\
       \affaddr{Yahoo Labs}\\
       \email{nanliu@yahoo-inc.com}
\and  % use '\and' if you need 'another row' of author names
% 4th. author
\alignauthor Suju Rajan\\
       \affaddr{Yahoo Labs}\\
       \email{suju@yahoo-inc.com}
% 5th. author
\alignauthor Max Welling
       \titlenote{M. Welling also has a partime position at the UC Irvine.}\\
       \affaddr{University of Amsterdam}\\
       \email{m.welling@uva.nl}
}
% There's nothing stopping you putting the seventh, eighth, etc.
% author on the opening page (as the 'third row') but we ask,
% for aesthetic reasons that you place these 'additional authors'
% in the \additional authors block, viz.
%\additionalauthors{Additional authors: John Smith (The Th{\o}rv{\"a}ld Group,
%email: {\texttt{jsmith@affiliation.org}}) and Julius P.~Kumquat
%(The Kumquat Consortium, email: {\texttt{jpkumquat@consortium.net}}).}
% Just remember to make sure that the TOTAL number of authors
% is the number that will appear on the first page PLUS the
% number that will appear in the \additionalauthors section.

\maketitle
\begin{abstract}
Despite having various attractive qualities such as high prediction accuracy and the ability to quantify uncertainty and avoid over-fitting, Bayesian Matrix Factorization has not been widely adopted because of the prohibitive cost of inference. In this paper, we propose a scalable distributed Bayesian matrix factorization algorithm using stochastic gradient MCMC. Our algorithm, based on Distributed Stochastic Gradient Langevin Dynamics, can not only match the prediction accuracy of standard MCMC methods like Gibbs sampling, but at the same time is as fast and simple as stochastic gradient descent. In our experiments, we show that our algorithm can achieve the same level of prediction accuracy as Gibbs sampling an order of magnitude faster. We also show that our method reduces the prediction error as fast as distributed stochastic gradient descent, achieving a 4.1\% improvement in RMSE for the Netflix dataset and an 1.8\% for the Yahoo music dataset. 

% while achieving a relative improvement in accuracy of 3.7\% on the Netflix movie ratings dataset and 1.1\% on the Yahoo music ratings dataset at the time when DSGD converges. Given more time this gap is increased up to 4.1\% and 1.8\% respectively.  
\end{abstract}

%\category{G.4}{Mathematical Software}{Parallel and vector implementations}
%A category including the fourth, optional field follows...
% \category{D.2.8}{Software Engineering}{Metrics}[complexity measures, performance measures]

%\terms{Algorithms, Experimentation, Performance}

%\keywords{Large-Scale, Distributed Matrix Factorization, MCMC, Stochastic Gradient, Bayesian Inference}

\section{Introduction}

Recommender systems have become a pervasive tool in industry to understand customers and their interests in products. Examples range between music recommendation (Pandora), book recommendation (Amazon), movie recommendation (Netflix), news recommendation (Yahoo) to partner recommendation (eHarmony). Recommender systems represent a personalized technology that can help filter at an individual level the enormous amounts of information that is available to us. Given the exponential growth of data, recommender systems are likely to play an increasingly important role to manage our information streams. 

During 2006-2011 Netflix \cite{netflixwww, netflix} ran a competition where teams around the world could develop and test new recommender technology on Netflix movie rating data. A few valuable lessons were learnt from that exercise. First, matrix factorization methods work very well compared to nearest neighbor type models. Second, averaging over many different models pays off in terms of prediction accuracy. One particularly effective model was Bayesian probabilistic matrix factorization (BPMF) \cite{ruslan08bpmf} where predictions are averaged over samples from the posterior distribution. Besides improved prediction accuracy, a full Bayesian analysis also comes with additional advantages such as probabilities over models, confidence intervals, robustness against overfitting, and incorporating prior knowledge and side-information \cite{adams10incorporating, PorteousAsuncionWelling10}. 

Unfortunately, since the number of user-product interactions can easily run into the billions, posterior inference is usually too expensive to be practical. Learning at that scale requires data and computation to be distributed over many machines and learning updates to only depend on small minibatches of the data. Effective distributed learning algorithms have been devised for alternating least squares (ALS) and stochastic gradient descent (SGD) \cite{gemulla11large, recht13parallel, zhuang13fast, christina12distributed, niu2011hogwild, hall10mapreduce, mcdonald10distributed, mann09effcient, zinkevich10psgd}. In particular, Distributed Stochastic Gradient Descent (DSGD) \cite{gemulla11large} has achieved a significant speed-up by assigning partitioned rating matrix blocks to workers and then by updating some ``orthogonal'' blocks in parallel using ``stratified'' SGD. DSGD outperformed other parallel SGD approaches such as PSGD \cite{hall10mapreduce, mcdonald10distributed} and ISGD \cite{mann09effcient, zinkevich10psgd} where SGD is applied also on some subsets of the ratings while synchronizing globally after each sub-epoch (PSGD) or once at the end of the training (ISGD). Unfortunately, so far it has proven difficult to apply these advances in distributed learning to posterior sampling in Bayesian matrix factorization models. For instance, for BPMF which requires $\cO((L+M)D^3)$ computation per iteration (with $L$ and $M$ are number of users and items, and $D$ is latent feature dimension), distributed computation has not nearly been as effective. 

In this paper, we propose a scalable and distributed Bayesian matrix factorization method which combines the predictive accuracy of Bayesian inference and the learning efficiency of stochastic gradient updates. To this end, we extend a recently developed MCMC method, called Stochastic Gradient Langevin Dynamics (SGLD) \cite{WellingTeh11}, so that the updates become efficient in the setting of distributed, large-scale matrix factorization. We adapt the SGLD updates to make them suitable for distributed learning on subsets of users and products (or blocks). Each worker manages only a small block of the rating matrix, and updates and communicates only a small subset of the parameters in a fully-asynchronous or weakly-synchronous fashion. Unlike distributed SGD where a single model is learnt, our method deploys multiple parallel chains over workers. Consequently, samples are collected at a much faster rate than ordinary MCMC and the multiple parallel chains can explore different modes of parameter space. Both features are reducing the variance and increasing the accuracies of our predictions.

In the experiments on the Netflix and Yahoo music datasets (the latter being one of the largest publicly available dataset for recommendation problems), we show that our method achieves the same level of accuracy as BPMF but an order of magnitude faster. Reversely, at almost the same efficiency as distributed SGD, our method achieves much better accuracy (4.1\% RMSE improvement for the Netflix dataset and 1.8\% for Yahoo music dataset). As such we believe that the method proposed in this paper is currently the most competitive matrix factorization method for industry scale problems.

%%%%%%%%%%%%%%%%%%%%%%%%%%%%%%
%%%%%%%%%%%%%%%%%%%%%%%%%%%%%%
\section{Preliminaries}
%%%%%%%%%%%%%%%%%%%%%%%%%%%%%%
\subsection{Bayesian Matrix Factorization}\label{sec:model_bpmf}
Suppose we have $L$ users and $M$ items. Our goal is to learn latent feature vectors $U_i, V_j \in \eR^D$ such that the rating $R_{ij}$ for item $j$ by user $i$ can be predicted as $R_{ij} \approx U_i \trns V_j$. We denote the entire rating matrix by $\bR \in \eR^{L \times M}$, and the latent feature matrices by $\bU \in \eR^{D\times L}$ and $\bV \in \eR^{D\times M}$, so that $\bR \approx \bU\trns \bV$.  Assuming a Gaussian error model, the likelihood of the parameters $\bU$ and $\bV$ can be written as:
\bea
p(\bR | \bU,\bV, \tau) \eqa \pd{i}{L}\pd{j}{M}\left[\cN(R_{ij} | U_i\trns V_j, \tau\inv) \right]^{I_{ij}}.\label{eqn:bpmf_lik}
\eea
% Considering that ratings usually have a specific (discrete) ranges, e.g. 1 to 5, learning $\tau$ is not so important part of the problem. Thus, throughout the paper, we fix $\tau = 1$ for simplicity. 
% ---I don't think the above argument is correct -- Anoop
where $I_{ij}$ is equal to 1 if user $i$ rated item $j$ and 0 otherwise. Throughout the paper, we fixed $\tau = 1$ for simplicity\footnote{All update equations are derived with $\tau=1$.}. Although, in theory, $\bU$ and $\bV$ can be learned by maximizing the likelihood above, this results in severe over-fitting because only a few ratings are known (i.e. $\bR$ is very sparse).  

Therefore, a Bayesian Probabilistic Matrix Factorization (BPMF) model was proposed to overcome this problem \cite{ruslan08bpmf}. In addition to controlling over-fitting through posterior averaging, BPMF also provides estimates of uncertainty through the posterior predictive distribution. The BPMF model as proposed in \cite{ruslan08bpmf} is as follows. We place priors on $\bU$ and $\bV$ as:
\bea
p(\bU|\mu_U, \La_U) \eqa \pd{i}{L} \cN(U_i | \mu_U, \La_U\inv),\\
p(\bV|\mu_V, \La_V)  \eqa \pd{j}{M} \cN(V_j | \mu_V, \La_V\inv).
\eea
We further place Gaussian-Wishart hyper-priors on the user and item hyperparameters $\Ta_U = \{\mu_U, \La_U\}$ and $\Ta_V = \{\mu_V, \La_V\}$:
\bea
p(\Ta_U | \Ta_0) \eqa \cN(\mu_U | \mu_0, (\bt_0\La_U)\inv) \cW(\La_U | W_0, \nu_0),\\
p(\Ta_V |\Ta_0)  \eqa \cN(\mu_V | \mu_0, (\bt_0\La_V)\inv) \cW(\La_V | W_0, \nu_0),
\eea
where $\nu_0$ is the number of degrees of freedom and $W_0$ is a $D\times D$ scale matrix. We collectively denote the parameters of the hyper-prior by $\Ta_0 = \{\mu_0, \bt_0, \nu_0, W_0\}$. 

At test time, the predictive distribution of an unknown rating $R_{ij}^*$ can be obtained by marginalizing over both model parameters $\bU, \bV$ and hyper-parameters $\Ta_U, \Ta_V$,
\bea
p(R_{ij}^*| \bR, \Ta_0) \eqa \int \int p(R_{ij}^* | U_i, V_j) p(\bU,\bV | \bR, \Ta_U, \Ta_V)\nn\\
&&p(\Ta_U, \Ta_V | \Ta_0) \upd\{\bU,\bV\} \upd\{\Ta_U, \Ta_V \}
%&\approx& \f{1}{T} \sm{t}{T} p(R_{ij}^* | U_i^\sml{t}, V_j^\sml{t}).
\eea
We can estimate this using a Monte Carlo approximation:
\bea
p(R_{ij}^*| \bR, \Ta_0)  &\approx& \f{1}{T} \sm{t}{T} p(R_{ij}^* | U_i^\sml{t}, V_{j}^\sml{t}).
\eea
where $\left\{ U_i^\sml{t}, V_{j}^\sml{t} \right\}$ is the $t$-th sample from the posterior distribution:
\bea
p(\bU,\bV,\Ta_U,\Ta_V | \bR, \Ta_0).
\eea
These samples can be generated using Gibbs sampling (Algorithm \ref{alg:bpmf}), since by conjugacy the conditional distributions of $U_i$ and $V_j$ are Gaussian, and those of $\Ta_U$ and $\Ta_V$ are Gaussian-Wishart. However, sampling from the conditional distribution of $U_i$ or $V_j$ involves $\cO(D^3) $ computations (for inverting a $D \times D$ precision matrix) and since this has to be done for each user and item, results in a total of $\cO((L+M)D^3)$ computations per iteration. Thus, BPMF using Gibbs sampling cannot scale up to real world recommender systems with millions of users and / or items.
%
%To sample from the posterior distribution, BPMF using Gibbs sampling alternates sampling from conditional distributions which are, by conjugacy, Gaussian distributions for $U_i$ and $V_j$ and Gaussian-Wishart distributions for $\Ta_U$ and $\Ta_V$. Therefore, we can directly collect independent samples from each of the conditionals. Alg. \ref{alg:bpmf} describes a pseudo-code of Gibbs sampling for BPMF. Obtaining the conditional posterior distribution for $U_i$ (or $V_j$), however, involves expensive $\cO(D^3) $ computations, i.e. inverting a $D \times D$ precision matrix $\La_i^* = \La_U + \al\sm{j}{M} I_{ij}[V_jV_j\trns]$ for each user $i$ and similar for each item $j$. As a result, a single iteration of the Gibbs sampling requires $\cO((L+M)D^3)$ computations. BPMF therefore has been restricted from large-scale problems where millions of users and items are prevalent.
%

\begin{algorithm}[t]
\caption{Gibbs Sampling for BPMF}\label{alg:bpmf}
\begin{algorithmic}[1]
\STATE Initialize model parameters $\bU^\sml{1}, \bV^\sml{1}$
\FOR {$t=1:T$}
	\STATE // Sample hyperparameters\\
	$\Ta_U^\sml{t} \sim p(\Ta_U | \bU^\sml{t}, \Ta_0)$, \quad      
	$\Ta_V^\sml{t} \sim p(\Ta_V  | \bV^\sml{t}, \Ta_0)$
	\FOR {$i=1:L$ {\bf in parallel}}
%		\STATE Sample user features\\
		\STATE $U_i^\sml{t+1} \sim p(U_i | \bR, \bV^\sml{t}, \Ta_U^\sml{t})$ // sample user features %= \cN(U_i | \mu_i^*, [\La_i^*]\inv)$
	\ENDFOR
	\FOR {$j=1:M$ {\bf in parallel}}
%		\STATE sample item features\\
		\STATE $V_j^\sml{t+1} \sim p(V_j | \bR, \bU^\sml{t}, \Ta_V^\sml{t})$ // sample item features%= \cN(V_j | \mu_j^*, [\La_j^*]\inv)$
	\ENDFOR
\ENDFOR
\end{algorithmic}
\end{algorithm}

Although it is possible to parallelize BPMF using MapReduce style synchronous global updates, the cubic order complexity still limits its applicability to small $D$. Also, we require a large number of workers to effectively distribute the $L+M$ cubic-order computations. Furthermore, since running the Gibbs sampler from scratch is too expensive, a separate SGD optimizer is usually deployed to reach near the Maximum-a-Posteriori (MAP) state before starting the Gibbs sampler. However, running two different large-scale distributed algorithms, each of which requires different optimal settings for the distribution of data and parameters, as well as cluster architectures, adds another considerable level of complexity.

%%%%%%%%%%%%%%%%%%%%%%%%%%%%%%%
\subsection{Stochastic Gradient Langevin Dynamics}

Assume we have a dataset of $N$ i.i.d. data points, denoted by $\cX = \{x_n\}_{n=1}^N$, which we model using a distribution $p(x|\ta)$ parameterized by $\ta \in \eR^D$. We choose a prior distribution $p(\ta)$ and our goal is to sample from the posterior distribution $p(\ta|\cX) \propto p(\cX|\ta)p(\ta)$ using MCMC. 

One way of obtaining efficient MCMC proposals is to use the gradient of the target density \cite{rossky78mala, Duane87, Neal11, GirolamiCalderhead10}, e.g. Langevin Dynamics  \cite{rossky78mala} is an MCMC algorithm which proposes candidate states according to:
\bea
\ta_{t+1} \law \ta_t + \f{\ep_t}{2} \left\{ \grad_{\ta_t} \log p(\ta_t) + \sum_{x\in \cX} g(\ta;x) \right\}  + \nu_t \nonumber \\ \text{where}~~\nu_t \sim \cN{0, \ep_t I)}
\eea
In the above, $\ep_t$ is the step size and  $g(\ta;x) = \nabla_\ta \log p( x | \ta)$ is the score. A Metropolis-Hastings (MH) test is then used to decide whether to accept or reject the proposal. The gradient information allows the Langevin algorithm to make proposals to high density regions and therefore have a high probability of acceptance. However, in large-scale problems where $N=|\cX|$ can be very large, the $\cO(N)$ computations per update, required for computing the gradient as well as for the MH test, is infeasible.

Stochastic Gradient Langevin Dynamics (SGLD)  \cite{WellingTeh11} 
is the first in a line of recently developed approximate MCMC algorithms \cite{AhnKorattikaraWelling12, PatTeh13sgrld, tianqi14sghmc, dingfang14sgnht, AhnShaWel14} that try to address this issue using noisy gradients that can be cheaply computed from a mini-batch of $n \ll N$ data points. SGLD uses the following update rule:
\bea
\ta_{t+1} \law \ta_t + \f{\ep_t}{2} \left\{ \grad_{\ta_t} \log p(\ta_t) + N\barg(\ta_t;\cM_{t})\right\}  + \nu_t \label{eqn:sgld}.
\eea
Here $\barg(\ta_t;\cM_{t}) = \f{1}{n}\sum_{x\in \cM_t} g(x;\ta_t)$, the mean score computed from a mini-batch $\cM_t$. SGLD converges to the true posterior distribution if the step size is annealed to zero at a rate that satisfies the following conditions:
\bea
\sum_{t=1}^\infty \ep_t = \infty, \hspace{10mm}\sum_{t=1}^\infty \ep_t^2 < \infty.\label{eqn:step_cond}
\eea
SGLD does not use accept-reject tests because the acceptance rate tends to one as the step size goes to zero. Therefore, unlike traditional MCMC algorithms which require $\cO(N)$ computations per iteration, SGLD requires only $\cO(n)$ computations. 

More generally, it is valid to replace $\barg(\ta_t;\cM_{t})$ in eqn. \ref{eqn:sgld} with any estimator $f(\ta,Z;X)$ that satisfies the following conditions: (i) it is an unbiased estimator of the true gradient i.e. $\eE_Z[f(\ta,Z;\cX)] = \barg(\ta;\cX)$ (ii) it has finite variance $\eV_Z[f(\ta,Z;\cX)]<\infty$. Here, the expectation and variance are w.r.t. the distribution  $p(Z;\cX)$ of the auxiliary random variable $Z$.
 
Distributed SGLD (DSGLD) \cite{AhnShaWel14} further extends the power of stochastic gradient MCMC using distributed computing. In DSGLD, the dataset is first partitioned and distributed to $S$ workers. Then, multiple chains collect samples in parallel by sampling for the length of a round (called a \textit{trajectory}) at a worker. After a round, each chain switches to a different worker. In \cite{AhnShaWel14}, it is shown that using the following valid SGLD update rule, we can collect samples from the posterior using the distributed datasets:
\bea
\ta_{t+1} \law \ta_{t} + \f{\ep_{t}}{2} \left\{ \grad_{\ta_{t}} \log p(\ta_{t}) + \f{N^\sml{s}}{v^\sml{s}} \barg(\ta_{t};\cM^\sml{s}_{t})\right\}  + \nu_{t}.\label{eqn:dsgld}
\eea
Here, $s$ is the index of the worker where a chain resides at iteration $t$, $N^\sml{s}$ is the size of the local dataset at worker $s$, and $v^\sml{s}$ is the normalized visiting rate to worker $s$ such that $\sum_{s}\nu^\sml{s} = 1$ and $\nu^\sml{s} \in (0,1)$. The mini-batch $\cM^\sml{s}_{t}$ is sampled only from the local dataset of worker $s$.

%%%%%%%%%%%%%%%%%%%%%%%%%%%%%%%%
%%%%%%%%%%%%%%%%%%%%%%%%%%%%%%%%
\section{Bayesian Matrix Factorization using SGLD}

%%%%%%%%%%%%%%%%%%%%%%
\subsection{Model}\label{sec:model}

We will now show how DSGLD can be used for BPMF. Instead of the model described in Section \ref{sec:model_bpmf}, we will use a slightly simplified model \cite{andriy07pmf,tianqi14sghmc}. We use the same likelihood as in eqn. \ref{eqn:bpmf_lik}, but choose simpler priors:
\bea
p(\bU|\La_U) \eqa \pd{i}{L} \cN(U_i | 0, \La_U\inv),\\
p(\bV|\La_V)  \eqa \pd{j}{M} \cN(V_j | 0, \La_V\inv).
\eea
Here, $\La_U$ and $\La_V$ are $D$-dimension diagonal matrices whose $d$-th diagonal elements are $\la_{U_d}$ and $\la_{V_d}$ respectively. We also choose the following hyper-priors:
\bea
\la_{U_d}, \la_{V_d} &\sim& \text{Gamma}(\al_0, \bt_0).
\eea

We choose this simplified model because the proposed method benefits mainly from performing a large number of inexpensive updates (i.e. collecting many samples) per unit time rather than very expensive but high quality updates. The above model is well suited for this because each latent vector can be updated in linear ( $\cO(D)$ ) time. At the same time, we still benefit from the power of Bayesian inference through marginalization of the important regularization parameters $\La = \{\La_U, \La_V\}$ as well as $\bU$ and $\bV$. 

Although it is possible to apply our method to the model in Section \ref{sec:model_bpmf}, updating the full covariance matrix is more expensive ($\cO(D^2)$ time per update) and therefore requires more time to converge without significant gain in accuracy (as per our pilot experiments).

%%%%%%%%%%%%%%%%%%%%%%%%%%
\subsection{Inference}

In the following section, we first present our algorithm in a single machine setting and later extend it for distributed inference. We alternate between sampling from $p(\bU,\bV | \bR, \La)$ using SGLD and sampling from $p(\La | \bR, \bU, \bV)$ using Gibbs.  

%%%%%%%%%%%%%%%%%%%%%%%%%%
\subsubsection{Sampling $\bU$, $\bV$ | $\La$, $\bR$ using SGLD}

Since, usually only $N \ll M \times L$ ratings are observed, the rating matrix $\bR$ is stored using a sparse representation as $\cX = \{x_n = (p_n, q_n, r_n)\}_{n=1}^N$,  where each $x_n$ is a  (user, item, rating) tuple and $N$ is the number of observed ratings.  The gradient of the log-posterior w.r.t.\footnote{We derive only w.r.t. $U_i$. Update rules for other parameters can be obtained by the same procedure.} $U_i$ is:
\bea
G(\cX) = \sm{n}{N}g_n(U_i; \cX)  - \La_U U_{i}
\eea
where 
\bea
g_n(U_i; \cX) = \eI[p_n=i | \cX] (r_n - U_{p_n}\trns V_{q_n})V_{q_n}
\eea
Here $\eI[p_n = i| \cX]$ is an indicator function that equals 1 if the $n$-th tuple in $\cX$ pertains to user $i$ and 0 otherwise.  To use SGLD, we need an unbiased estimate of this gradient that can be computed cheaply from a mini-batch.
%The predicted rating is denoted by $y_n = U_{p_n}\trns V_{q_n} + a_{p_n} + b_{q_n}$.

One way to obtain this is by subsampling a mini-batch $\cM=\{(p_n,q_n,r_n)\}_{n=1}^m$ of $m$ tuples from $\cX$ and computing the following stochastic approximation of the gradient:
\bea
%G_1(\cM) =\f{N}{m}\sm{n}{m}g_n(U_i;\cM)  - \La_U U_{i}\\
G_1(\cM) =N\barg(U_i;\cM)  - \La_U U_{i}
\eea
where, $\barg(U_i;\cM) = \f{1}{m}\sm{n}{m}g_n(U_i;\cM)$. Note that the mini-batch is subsampled from the complete dataset $\cX$ and not just from the tuples associated with user $i$. The expectation of $G_1$ over all possible mini-batches is:
\bea
\eE_\cM [G_1(\cM)] \eqa \eE_\cM \left[N \barg(U_i;\cM)\right]  - \La_U U_{i} \nn\\
\eqa  \sm{n}{N}g_n(U_i;\cX)  - \La_U U_{i}\nn\\
\eqa G(\cX). \nn
\eea
Since $G_1$ is an unbiased estimator of the true gradient, we can use it for computing SGLD updates. However,  note that $G_1$ is non-zero even for users that are not in the mini-batch $\cM$, because of the prior gradient term $-\La_U U_i$. Therefore, we have to update the parameters for all  users in every iteration, which is very expensive.

If we were to update only the parameters of users who have ratings in the mini-batch $\cM$, the estimator can be written as:
\bea
G_2(\cM) = N \barg(U_i;\cM) - \eI[i \in \cM_p] \La_U U_{i}
\eea
where $\eI[i \in \cM_p]$ is equal to 1 if $\cM$ contains a tuple associated with user $i$ and 0 otherwise. However, $G_2$ is not an unbiased estimator of the true gradient:
\bea
\eE_\cM[G_2(\cM)] = \sm{n}{N}g_n(U_i;\cX) - h_{i*} \La_U U_i.
\eea
where $h_{i*} = \eE_\cM[\eI[i \in \cM_p]]$, i.e. the fraction of mini-batches that contains at least one tuple associated with user $i$ (among all possible mini-batches). If the mini-batches are sampled with replacement, we can compute this as:
\bea
h_{i *} = 1 - \left( 1 - \f{N_{i*}}{N}\right)^m
\label{eqn:biascorr}
\eea
where $N_{i*} = \sum_{n=1}^N \eI[p_n = i | \cX]$, the number of ratings by user $i$ in the complete dataset $\cX$. Thus, we can remove the bias in $G_2$ by multiplying the gradient of the prior term with $h_{i^*}\inv$ as follows:
\bea
G_3(\cM) = N \barg(U_i;\cM) -\eI[i \in \cM_p] h_{i*}\inv\La_U U_{i}.
\eea
$G_3$ is an unbiased estimator of the true gradient $G$ and is non-zero only for users that have at least one rating in $\cM$. Thus we need to update only a subset of user features in each iteration. The SGLD update rule (for users with ratings in $\cM_t$) is:
\bea
U_{i,t+1} \law U_{i,t} + \f{\ep_t}{2}\left\{N \barg(U_{i,t};\cM_t) - \f{\La_U U_{i,t}}{h_{i *}}\right\} + \nu_t 
\eea

%%%%%%%%%%%%%%%%%%%%%%%%%%
\subsubsection{Sampling $\La | \bU, \bV$}
We can easily sample from the conditional $p(\La | \bU, \bV)$, because by conjugacy: 
\bea
\la_{U_d} | \bU,\bV \sim \text{Gamma}\left(\al_0 + \f{L}{2}, \bt_0 + \ha \sm{i}{L}U_{di}^2\right),\label{eqn:gibbs_la_u}\\
\la_{V_d} | \bU,\bV \sim \text{Gamma}\left(\al_0 + \f{M}{2}, \bt_0 + \ha \sm{i}{M}V_{dj}^2\right)\label{eqn:gibbs_la_v}.
\eea
If this is computationally demanding, we can also consider updating $\La$ using SGLD or mini-batch Metropolis-Hastings \cite{korattikara14austerity, remi2014toward}.
%%%%%%%%%%%%%%%%%%%%%%%%%%%%%%%%
\subsection{Distributed Inference}

%{\bf Connection to DSGD} If $|\cM^\sml{s}| = 1$
%%%%%%%%%%%%%
\begin{figure}[t]
	\centering
	\vspace{0.0cm}
		\subfigure[square]{
		\includegraphics[width=0.0855\textwidth]{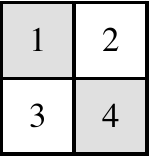}}
		\hspace{0.7cm}
		\subfigure[column]{
		\includegraphics[width=0.080\textwidth]{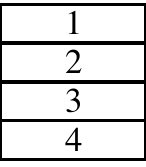}}
		\hspace{0.7cm}
		\subfigure[hybrid]{
		\includegraphics[width=0.080\textwidth]{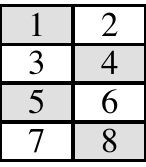}}
\vspace{0.0cm}
\caption{Block split schemes.} 
\vspace{0.0cm}
\label{fig:block_split}
\end{figure}

%%%%%%%%%%%%%%%%%%%%%%%%%%%%%%%
For distributed inference, we partition the rating matrix $\bR$ into a number of blocks.  Fig. \ref{fig:block_split} shows a few different ways of partitioning $\bR$. Two blocks are said to be \textit{orthogonal} to each other if the users and items in one block do not appear in the other block. A set of two or more mutually orthogonal blocks is called an \textit{orthogonal block group} (or simply, orthogonal group). For example, the two gray-colored blocks (1 and 4) in Fig. \ref{fig:block_split} (a) are orthogonal to each other and thus form an orthogonal group. In Fig. \ref{fig:block_split} (b), the blocks are not orthogonal because all columns are shared. In this case, we say that each block by itself is an orthogonal group. 

The blocks are then distributed to workers in such a way that all blocks are assigned and a worker has at least one block. In the following, we assume for simplicity that each worker is a single-core machine. However, it is easy to generalize our algorithm to take advantage of multi-core (or threads) workers with shared memory support. 

We will now describe our distributed algorithm for BPMF. First, imagine that there is only one Markov chain $c$ (but the dataset is distributed across multiple workers). A central parameter server holds the \emph{global} parameters $\bU^{c}$ and $\bV^{c}$ of chain $c$. Since $\La$ depends only on $\bU^{c}$ and $\bV^{c}$, it is easy to update $\La$ at the parameter server using Gibbs as per Eqns. \ref{eqn:gibbs_la_v} and \ref{eqn:gibbs_la_u}. Thus, we will focus on the DSGLD part of the chain that samples from $p(\bU,\bV|\bR, \La)$. 

Each sampling round consists of the following steps: (1) The parameter server picks a block $s$ via a block-scheduler and sends the corresponding sub-parameter $\bU^\sml{c,s}$ and $\bV^\sml{c,s}$ to the block's worker. (2) The worker updates the sub-parameter by running DSGLD (see section \ref{sec:dsgld_updates} for update equations) for a number of iterations using its local block of ratings. (3)  The worker sends the final sub-parameter state back to the parameter server. (4) The parameter server updates its global copy to the new sub-parameter state.

Thus, the Markov chain jumps among the distributed blocks through the corresponding workers and updates the sub-parameters associated with the block chosen in each round. Since each iteration of local DSGLD updates requires only a mini-batch of data, sampling is very fast. Also, communication overhead is low because a) the multiple local updates (iterations) performed within a round do not require any communication b) only a small sub-parameter associated with a specific block is transferred in each round. There are two levels of parallelization that we use to further speed up sampling.

% \vspace{-0.0mm}
%\begin{enumerate}[noitemsep]
% \benum

{\bf 1. Parallel updates within a chain: }. A chain can update sub-parameters $\bU^\sml{c,s_1}$ and $\bU^\sml{c,s_2}$ in parallel if the blocks $s_1$ and $s_2$ are orthogonal to each other. For example, in Fig. \ref{fig:block_split} (a), updating block 1 \textit{and then} block 4 produces the same result as updating both in parallel. This makes the algorithm progress faster in terms of number of updated parameters per round. The actual performance improvement is dependent on the size of the orthogonal group. For instance, with a $4 \times 4$ split, the algorithm will update the parameters faster than with a $2 \times 2$ split because more parameter blocks can be updated in parallel. However, updates in smaller blocks can be noisier, because the gradients computed from smaller blocks will have higher variance. Therefore, at some point the loss in performance caused by noisier updates on small blocks can exceed the gain obtained by faster updating of the parameters.

{\bf 2. Multiple parallel chains: } We can run as many chains in parallel as we like, subject to only computational resource constraints. Each chain can update its parameters in parallel independent of other chains. Hence, the chains are asynchronous in the sense that the status of a chain does not block other chains unless the chains conflict for computation resources. For the split in Fig. \ref{fig:block_split} (a), one chain can update using the gray block group while another chain is using the white block group. Or both chains can use the same block if we assume a shared memory multi-threaded implementation. By running multiple chains in parallel, we effectively multiply the number of collected samples by the number of parallel chains. Since the variance of an MCMC estimator is inversely proportional to the number of samples, fast sample generation will compensate for the low mixing rate of SGLD. Also, by initializing the different chains in different places of parameter space, we can explore multiple local minima. This is especially important for large-scale high dimensional problems where the time budget is usually not enough for a single chain to mix between different local minima. 
% \eenum
% \vspace{-0.0mm}

%%%%%%%%%%%%%
\begin{figure}[t]
	\centering
	\vspace{-0.0cm}
	\hspace{-0.0cm}
		\includegraphics[width=0.480\textwidth]{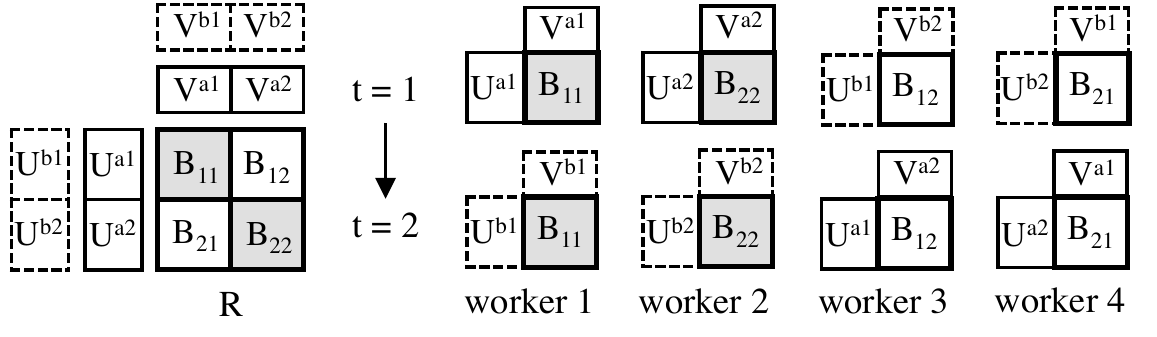}
%\vspace{-0.5cm}
\caption{An example illustration. On the left, a matrix $\bR$ is partitioned into $2 \times 2$ blocks, $\bB_{11}, \cdots, \bB_{22}$. There are two orthogonal groups (the gray $(\bB_{11}, \bB_{22})$ group and the white $(\bB_{12}, \bB_{21})$ group). We run two independent chains, chain $a$ with parameters $\bU^a$ and $\bV^a$ (solid-line rectangles) and chain $b$, with parameters $\bU^b$ and $\bV^b$ (dotted-line rectangles). Given four workers, we assign a block to each worker.  At round $t=1$, chain $a$ updates using the gray orthogonal group and chain $b$ using the white orthogonal group. Note that the entire $\bU$ and $\bV$ matrices of both chains are updated in this single round. In the next round, the chains are assigned to the next orthogonal groups by the block-scheduler.}
\vspace{-0.0cm}
\label{fig:illust}
\end{figure}
%%%%%%%%%%%%%

\begin{algorithm}[t]
\caption{DSGLD at parameter server}\label{alg:paraserver}
\begin{algorithmic}[1]
\STATE Initialize model parameters of each chain $\{\bU_1^{c}, \bV_1^{c}, \La_1^{c}\}_{c=1}^C$, step sizes $\{\ep_{t}\}$
\FOR{each chain $c$ \textbf{\textit{parallel}}}	
	\FOR {$t$=1:max\_iter}
		\STATE $\cB_c$ $\law$ \textsc{get\_ortho\_block\_group($c,t$)}
		\FOR{worker $s \in \cB_c$}
			\STATE $\bU_{t+1}^\sml{c,s},\bV_{t+1}^\sml{c,s} \law$ \textsc{wkr\_round($\bU_{t}^\sml{c,s},\bV_{t}^\sml{c,s}, \La_t^\sml{c}, \ep_{t}$)}
		\ENDFOR
		\IF {not burn-in}
		\STATE Store $\bU_{t+1}^\sml{c},\bV_{t+1}^\sml{c}$ as a sample of chain $c$
		\STATE Sample $\La_{t+1}^\sml{c} | \bU_{t+1}^\sml{c},\bV_{t+1}^\sml{c}$ using Eqn. \eqref{eqn:gibbs_la_u} and \eqref{eqn:gibbs_la_v}
		\ENDIF
	\ENDFOR
\ENDFOR
\end{algorithmic}
\end{algorithm}

\begin{algorithm}[t]
\caption{DSGLD at worker $s$}\label{alg:worker}
\begin{algorithmic}[1]
\STATE Initialize $\bar{h}_{i*}, \bar{h}_{* j}$, round length $\ga$, mini-batch size $m$
\FUNCTION {\textsc{wkr\_round}$(\bU^\sml{c,s},\bV^\sml{c,s}, \La^\sml{c}, \ep_{t})$}
\FOR {$t=1:\ga$}
 	\STATE Sample a mini-batch $\cM_{t}$ from $\cX^\sml{s}$
	\FOR {each user $i$ and item $j$ in $\cM_{t}$ \textbf{\textit{parallel}}}
 	\STATE Update $U_i, V_j$ using Eqn. \eqref{eqn:dsgld_dbmf_u} and \eqref{eqn:dsgld_dbmf_v}
% 	\STATE Update $$ by Eqn. \eqref{eqn:dsgld_dbmf_v}
	\ENDFOR
%	\FOR {each item $j \in \cM_{p,t}$ \textbf{\textit{parallel}}}
%	\ENDFOR	
\ENDFOR
\STATE Send updated $\bU^\sml{c,s}$ and $\bV^\sml{c,s}$ to the parameter server
\ENDFUNCTION
\end{algorithmic}
\end{algorithm}

An illustration of these ideas is given in Fig. \ref{fig:illust}. Algorithms \ref{alg:paraserver} and \ref{alg:worker} describe the operations at the parameter server and workers respectively. 

A proper block splitting scheme can be chosen according to the characteristics of the problem and available resources. In other words, we can trade-off \textit{within-chain} parallelization and \textit{between-chain} parallelization. For example, given $S$ workers, by using a squared split as in Fig. \ref{fig:block_split} (a), we can run $\sqrt{S}$ chains in parallel where each chain updates $\sqrt{S}$ blocks in parallel. This way we maximize the within-chain parallelism. On the other hand, by reducing the size of orthogonal groups, we can decrease the within-chain parallelism in order to increase the between-chain parallelization, i.e. number of parallel chains. At an extreme of this approach, we can let each block become an orthogonal group by itself as in Fig. \ref{fig:block_split} (b) and run $S$ independent chains in parallel. Note that in this case, we can choose not only the column splitting but any splitting scheme. Our experiment results suggest to maximize the within-chain parallelism as the dataset size increases. For smaller datasets, we may benefit more from the generalization performance of a large number of parallel chains than from a smaller number of chains using the block orthogonality.

% choose to not do any parallelization within each chain in order to increase the number of parallel chains, and just run $S$ independent chains in parallel. In this case, we can choose any splitting scheme.

\subsubsection{Distributed SGLD}\label{sec:dsgld_updates}

Since $\cX$ (the sparse representation of $\bR$) is partitioned into $S$ blocks $\cX^\sml{1}, \dots, \cX^\sml{S}$, each worker uses only one of the $\cX^\sml{s}$ for computing updates.  Thus, we need to modify the bias correctors in Eqn. \eqref{eqn:biascorr} so that the gradient estimator remains unbiased under this constraint. If we assume  $\cup_{s=1}^S \cX^\sml{s} = \cX$ and $\cap_{s=1}^S \cX^\sml{s} = \emptyset$, and that worker $s$ is visited with normalized frequency $v^\sml{s}$, the correction factors for users and items can be shown to be, respectively:
\bea
\bar{h}_{i *} = \sm{s}{S}v^\sml{s} h_{i*}^\sml{s}, \quad\quad \bar{h}_{* j} = \sm{s}{S}v^\sml{s} h_{* j}^\sml{s}
\eea
where:
\bea
h_{i*}^\sml{s} = 1 - \left( 1 - \f{N^\sml{s}_{i*}}{N^\sml{s}}\right)^m, \hspace{2mm} h_{* j}^\sml{s} = 1 - \left( 1 - \f{N^\sml{s}_{*j}}{N^\sml{s}}\right)^m
\eea
here $N^\sml{s} = |\cX^\sml{s}|$, the total number of ratings in $s$, and 
\bea
N^\sml{s}_{i*} = \sum_{n=1}^{N^\sml{s}} \eI[p_n = i | \cX^\sml{s}], \quad
N^\sml{s}_{*j} = \sum_{n=1}^{N^\sml{s}} \eI[q_n = j | \cX^\sml{s}].
\eea
i.e. the number of ratings by user $i$ and of item $j$ respectively in $s$. Therefore, the local DSGLD update rule using block $\cX^\sml{s}$ is: 
\bea
U_{i,t+1} \law U_{i,t} + \f{\ep_t}{2}\left\{\f{N^\sml{s}}{v^\sml{s}}
\barg(U_{i,t} ;\cM_{t}^\sml{s}) - \f{\La_UU_{i,t}}{\bar{h}_{i*}} \right\} + \nu_{t} && \label{eqn:dsgld_dbmf_u}\\
V_{j,t+1} \law V_{j,t} + \f{\ep_{t}}{2}\left\{\f{N^\sml{s}}{v^\sml{s}}\barg(V_{j,t};\cM_{t}^\sml{s}) - \f{\La_VV_{j,t}}{\bar{h}_{* j}} \right\} + \nu_{t}.&& \label{eqn:dsgld_dbmf_v}
\eea
The above rule updates only the sub-parameter associated with block $s$ using only rating tuples in $s$.

%%%%%%%%%%%%%%%%%%%%%%%%
%%%%%%%%%%%%%%%%%%%%%%%%
\section{Experiments}
%%%%%%%%%%%%%%%%%%%%%%%%
\subsection{Algorithms and Models}
\begin{table}[h]
	\centering
    	\begin{tabular}{ c | c  c  }
%    	\hline
     	     		  & Optimization   	&  MCMC \\ \hline %\hline
    	Single Machine & SGD            &  SGLD, Gibbs      \\ %\hline
	Distributed    &DSGD           	&  DSGLD   \\ 
%	\hline
    	\end{tabular}
	\caption{Algorithms.}	\label{tb:algos}	
\end{table}

We compared five algorithms: SGD, DSGD, SGLD, DSGLD, and Gibbs sampling. As shown in Table \ref{tb:algos}, each algorithm can be classified based on whether it is running on a single machine or a distributed architecture, and also based on whether it is an optimization or MCMC algorithm. Since Gibbs sampling was very slow, we update user/item features in parallel (as suggested in \cite{ruslan08bpmf}) using multiple cores of a single machine. Thus, by Gibbs sampling we will mean the parallelized (but not distributed) version from now on. 

For DSGLD, we tested two block-splitting schemes. Given $S$ workers, DSGLD-S (`S' stands for square) partitions $\bR$ into $\sqrt{S} \times \sqrt{S}$ blocks as in Fig. \ref{fig:block_split} (a), i.e. DSGLD-S tries to maximize  the within-chain parallelism by using as many orthogonal blocks as possible. We run $\sqrt{S}$ parallel chains, where each chain updates $\sqrt{S}$ sub-parameter blocks in parallel using $\sqrt{S}$ workers. Therefore, all chains can update all parameter at every round. The second splitting scheme, called DSGLD-C (`C' stands for column blocks) divides $\bR$ into $S$ blocks as shown in Fig. \ref{fig:block_split}(b). We split $\bR$ along the rows because in our experiments we have many more users than items. The blocks in DSGLD-C are not orthogonal because all columns are shared, so we just run $S$ independent parallel chains.

For Gibbs sampling, we use the original BPMF model\footnote{Using the simplified model does not reduce the computation complexity of the Gibbs sampling.} described in section \ref{sec:model_bpmf}. For the other algorithms, we slightly extend the model described in section \ref{sec:model} (as in \cite{tianqi14sghmc, koren09mf}). The extension includes user and item specific bias terms $a_i$ and $b_j$ respectively so that the predictions are modeled as:
\bea
R_{ij} \approx U_i\trns V_j + a_i + b_j
\eea
We use the following priors and hyper-priors for $a_i$ and $b_j$:
\bea
&a_i \sim \cN(0, \la_{a}\inv), \quad b_j \sim \cN(0, \la_{b}\inv),&\nn\\
&\la_{a}, \la_{b} \sim \text{Gamma}(\al_0, \bt_0).&\nn
\eea
For $\bU$ and $\bV$, we use the same priors and hyper-priors as described in Section \ref{sec:model}. Note that, in the new model, we have to sample $a_i, b_j, \la_a, \la_b$ in addition to $\bU, \bV, \La_U, \La_V$. The DSGLD update rules for $a_i$ and $b_j$ are: 
\bea
a_{i,t+1} \law a_{i,t} + \f{\ep_t}{2}\left\{\f{N^\sml{s}}{v^\sml{s}}
\barg(a_{i,t} ;\cM_{t}^\sml{s}) - \f{\la_a a_{i,t}}{\bar{h}_{i*}} \right\} + \nu_{t} && \label{eqn:dsgld_dbmf_a}\\
b_{j,t+1} \law b_{j,t} + \f{\ep_{t}}{2}\left\{\f{N^\sml{s}}{v^\sml{s}}\barg(b_{j,t};\cM_{t}^\sml{s}) - \f{\la_b b_{j,t}}{\bar{h}_{* j}} \right\} + \nu_{t}.&& \label{eqn:dsgld_dbmf_b}
\eea
The main goal of our experiments is to answer the following questions: 
\bitem
\item \textit{Accuracy}: How does DSGLD compare to other methods in terms of prediction RMSE?
\item \textit{Speed}: How fast can DSGLD achieve the RMSE obtained by 1) optimization algorithms (SGD, DSGLD) 2) Gibbs sampling?
\item \textit{Factors which affect the above}: The number of workers, number of chains, block splitting schemes and the latent factor dimension. 
\eitem

%%%%%%%%%%%%%%%%%%%%%%%%
\subsection{Setup}

\begin{table}[h]
	\centering
    \begin{tabular}{ c | c  c  c }
%    	\hline
     Dataset	& \# users  &  \# items  & \#  ratings \\ \hline
%	\hline
    	Netflix 	& 480K       & 18K      &  100M\\ 
%	\hline
	Yahoo      & 1.8M        & 136K	     & 700M\\ 
%	\hline
    	\end{tabular}
	\caption{Datasets.}	\label{tb:datasets}	
\end{table}

We compare all 5 algorithms on two large datasets,  Netflix movie ratings \cite{netflix} and Yahoo music ratings \cite{yahoomusic_r2} (details in Table \ref{tb:datasets}). To the best of our knowledge, the Yahoo dataset was one of the largest publicly available datasets when we performed the experiments. Note that the Yahoo dataset we use here is different from the one used in the KDD'11 Cup \cite{dror12kddcup11} (which has $\sim$250M music ratings and is often referred to by the same name). For the Netflix dataset, we use 80\% of the ratings for training and the remaining 20\% for testing as in \cite{dingfang14sgnht}. For the Yahoo dataset, the memory footprint was around 17GB for the train and test ratings, and around 1GB for $\bU$ and $\bV$ with $D=60$ in our 64-bit float based implementation. The memory footprint of the Netflix dataset was relatively small.

We used Julia \cite{bez14julia} to configure the cluster and execute the core routines of the algorithms. The core routines were implemented in \verb!C! for high performance. For distributed computing, we used Amazon EC2 instances \cite{amazonec2} of type ``r3" which were equipped with Intel Xeon 2.5 GHz CPUs and had memory configurable up to 244GB. Although the instances had multiple cores, we restricted all algorithms, except Gibbs sampling, to run on a single-core. For Gibbs sampling, we used a 12-core machine with the same CPU speed. All algorithms were implemented as an in-memory execution model and thus no disk I/O overheads were considered.

%%%%%%%%%
\begin{figure*}[t]
	\centering
	\vspace{-0.0cm}
%		\hspace{-0.4cm}
		\includegraphics[width=0.35\textwidth]{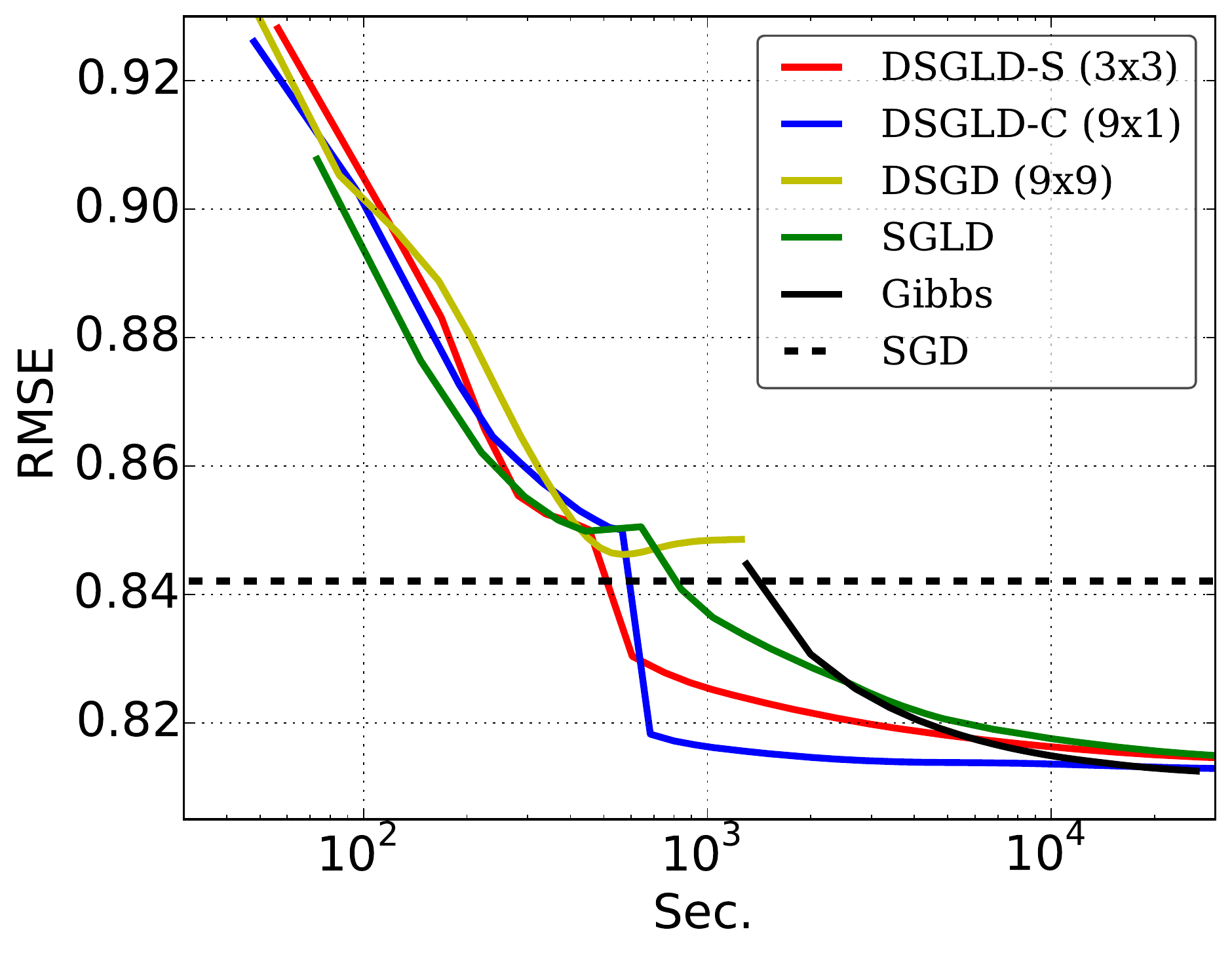}
		\hspace{2.5cm}
		\includegraphics[width=0.355\textwidth]{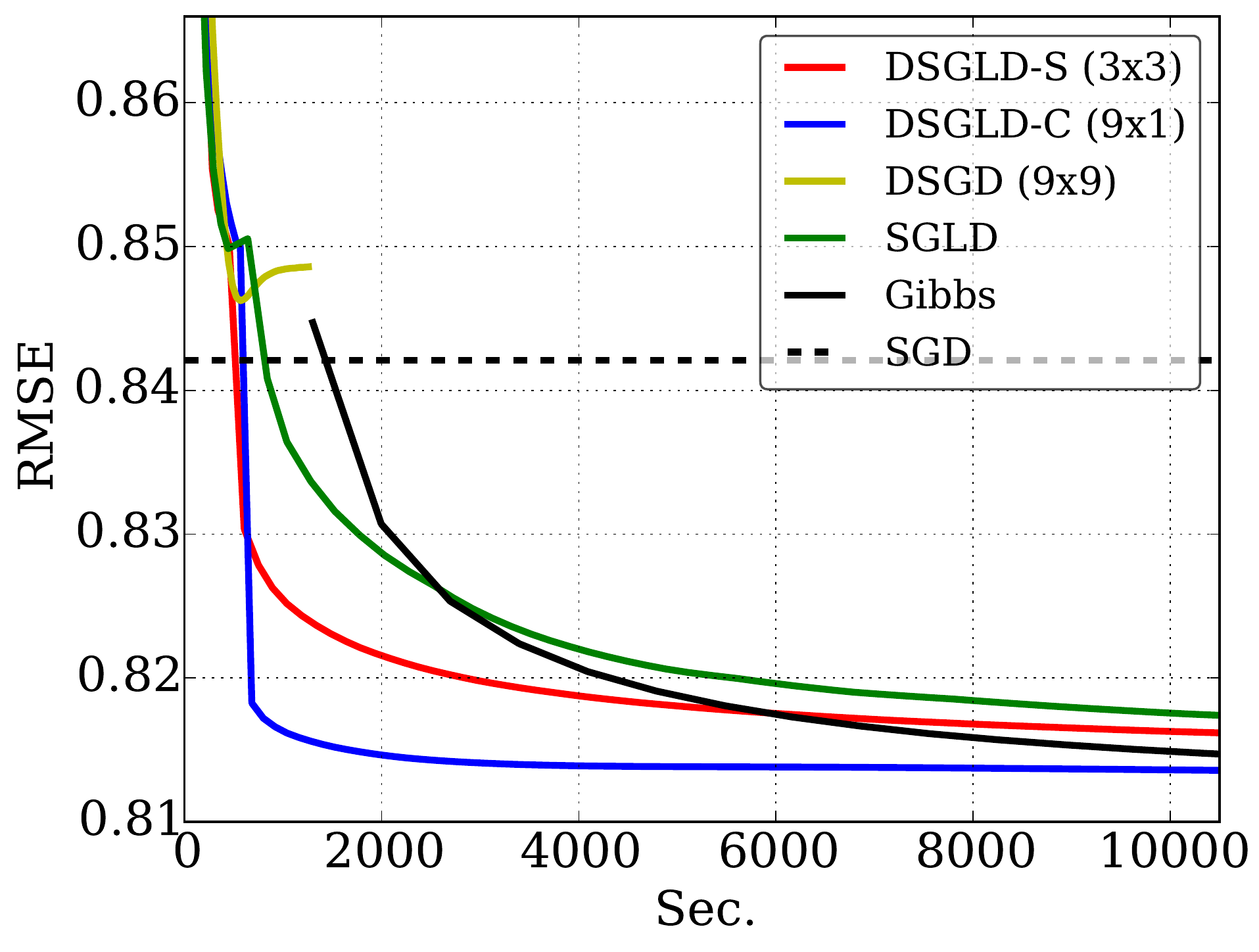}
\vspace{-0.3cm}
\caption{Netflix dataset ($D=30$).}
\label{fig:netflix_d30}
\end{figure*}
%%%%%%%%%

%%%%%%%%%
\begin{figure*}[t]
	\centering
	\vspace{-0.0cm}
		\hspace{-0.1cm}
		\includegraphics[width=0.35\textwidth]{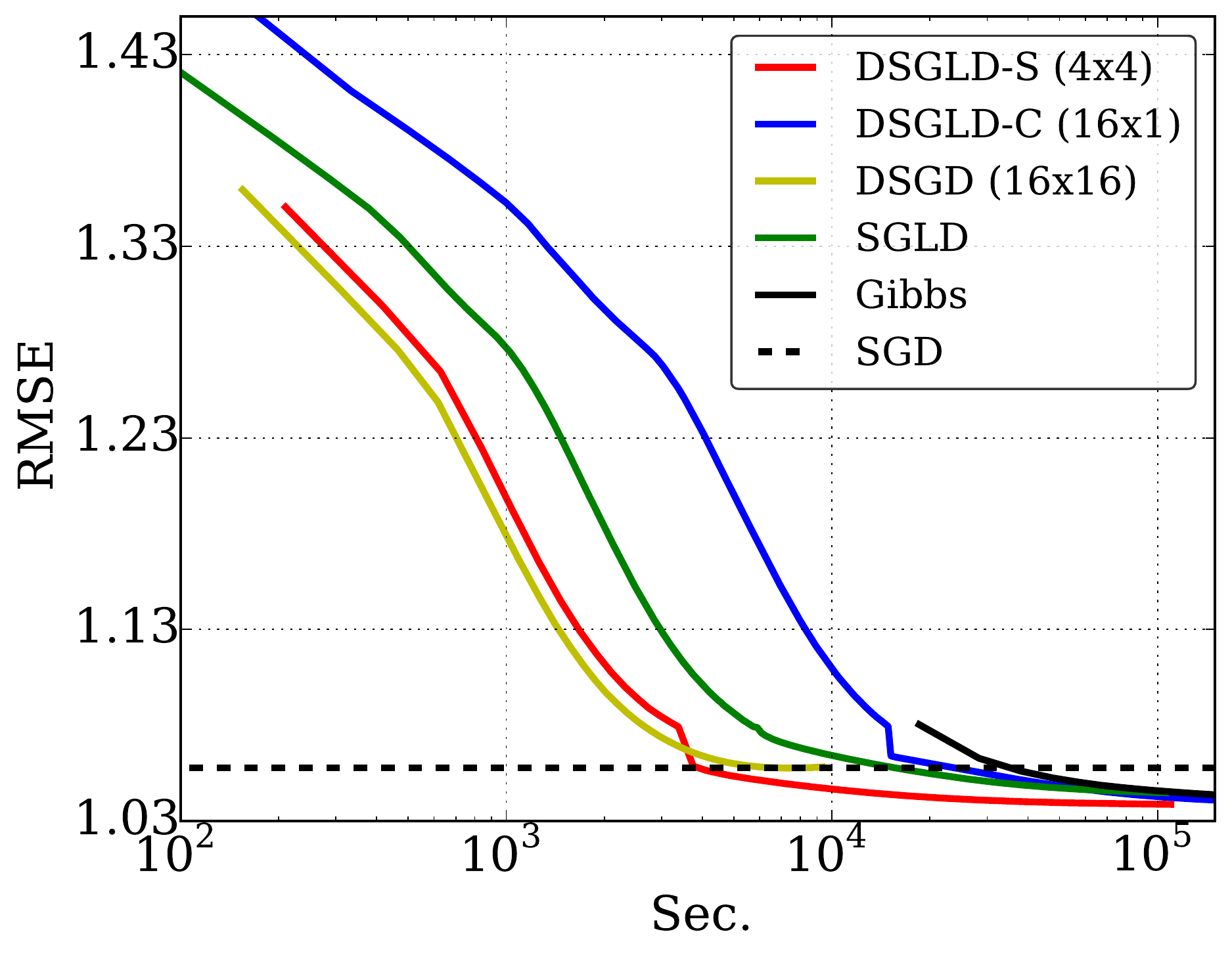}
		\hspace{2.5cm}
		\includegraphics[width=0.355\textwidth]{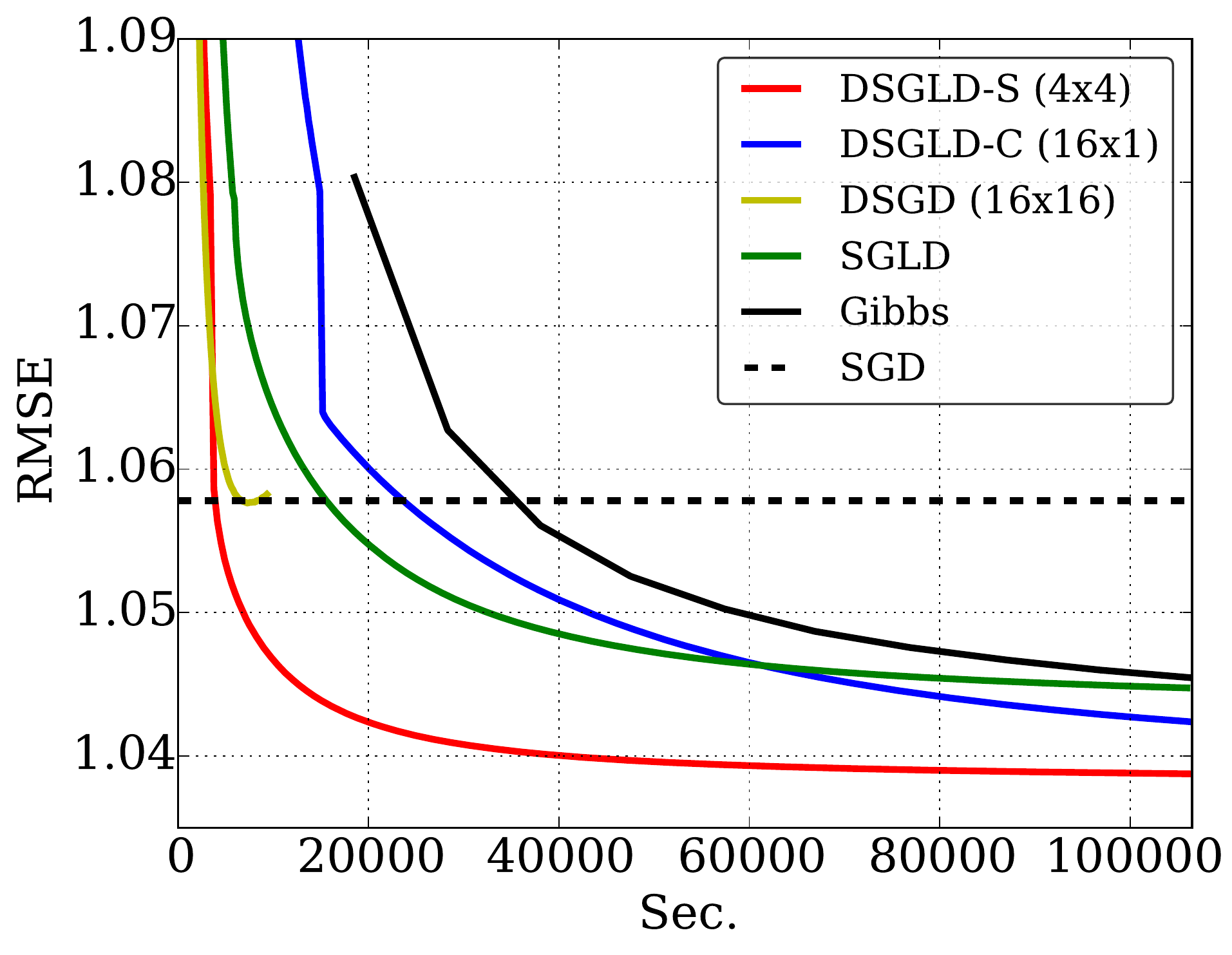}
\vspace{-0.3cm}
\caption{Yahoo Music Rating dataset ($D=30$).}
\vspace{-0.0cm}
\label{fig:yahoo_d30}
\end{figure*}
%%%%%%%%%

We annealed the step size according to the schedule $\ep_t = \ep_0 (1 + t/\ka)^{-\ga}$, (as in  \cite{AhnKorattikaraWelling12, PatTeh13sgrld}) which satisfies the convergence conditions in Eqn. \eqref{eqn:step_cond}. We found $\ka$, which controls the decay rate, over the range $\ka$ = $[10,50,100,500,1000,1500]$. The initial step size $\ep_0$ was also selected from $[$9e-6$, $1e-6$]$ for Netflix and $[$3e-6$, $8e-7$]$ for Yahoo. More detailed settings are given in the Appendix. We decreased the stepsize after every \textit{round} which we set to 50 updates. We used $\ga = 0.51$ in all experiments.

We set the hyperparameters $\tau=2.0$ and $\al_0 = 1.0$ for all experiments. We used $\bt_0=1.0$ for all algorithms except SGLD and DSGLD. For SGLD and DSGLD, the scale of the prior gradients sometimes became large due to multiplication by the bias correctors $1/h_{i *}$ and $1/h_{* j}$. In this case, instead of increasing the mini-batch size to reduce the scale of the correctors, we used a more appropriate scale parameter for the  Gamma prior distribution ($\bt_0 = 300$), to stabilize the scale of precisions sampled from the posterior Gamma distribution. 

Mini-batch sizes were set to 50K data points for Netflix and 100K for Yahoo. The initial values for the precisions $\La$ 
% $\la_{U_d}, \la_{V_d}, \la_a, \la_b$ 
were all chosen to be $2.0$ after testing over a range $[10, 5, 2, 1, 0.1, 0.01]$. In SGLD and DSGLD, the precision parameters were sampled every 50 rounds after burn-in. We discarded (burned) samples until the RMSE reached 0.85 for Netflix and 1.08 for Yahoo. For DSGLD, which deploys multiple chains, we used the arithmetic mean of the RMSE of all chains to determine whether burn-in has completed. We set the thinning interval to 10 rounds, i.e. we use only every $10^{th}$ sample to compute the average prediction. The Gibbs sampler in our experiments was initialized near a MAP state which we found using SGD during burn-in.

Running DSGLD requires a block scheduler (line 4 in Algorithm \ref{alg:paraserver}) that determines which blocks (workers) are used by each chain in a round. In our experiments, the blocks and the orthogonal groups were chosen beforehand and were assigned to chains deterministically using a cyclic-shift (rotation) at every round with equal visiting frequency. This scheduling policy is illustrated in Fig. \ref{fig:illust}.

%%We fixed the orthogonal groups before the training begins, and assigned a group to a chain while deterministically rotating (cyclic-shift) the orthogonal groups by a specific order with equal visiting frequency - doesn't this apply to non orthogonal blocks as well? - Anoop

%%%%%%%%%%%%%%%%%%%%%%%%%%%%%
\subsection{Results}
\subsubsection{Convergence and wall-clock time}
We first compare the RMSE of the algorithms as a function of computational time. In this experiment, we set $D$=30 for both datasets and used 9 workers for Netflix and 16 workers for Yahoo. Given $S$ workers, we used a $\sqrt{S} \times \sqrt{S}$ block-split for DSGLD-S, $S \times 1$ split for DSGLD-C and $S \times S$ split for DSGD. The total runtime was set to 50K seconds ($\approx$14 hours) for Netflix and 100K seconds ($\approx$27 hours) for Yahoo. In both Figs. \ref{fig:netflix_d30} and \ref{fig:yahoo_d30},  the x-axis is in log-scale for the figure on the left and in linear-scale for the figure on the right. 

In Fig. \ref{fig:netflix_d30}, we show results on the Netflix dataset (which is smaller than the Yahoo dataset). We see that in the early (burn-in) stage, all algorithms except Gibbs reduce error at a similar rate. Even though DSGLD-S and DSGD uses block orthogonality to update the sub-parameters of a chain in parallel, because of communication overheads, the gain in speed-up is not enough to outperform a non-distributed algorithm like SGLD which is able to reduce the error at a similar rate (because the dataset size is not very large) without any communication overhead. Note that because there are many chains for DSGLD, we plot the RMSE from only one chain during burn-in. The variance of RMSE across the chains was small during burn-in.

When the burn-in phase ends at around 500 - 700 seconds, MCMC algorithms (SGLD, DSGLD, and Gibbs) begin to collect samples and average their predictions over the samples, while DSGD does not and begins to overfit. Interestingly, at this point, we see a remarkably steep decrease in error for both DSGLD-S and DSGLD-C. In particular, we see the largest decrease for DSGLD-C which deploys 9 independent chains (whereas DSGLD-S uses 3 chains). Note that this is not solely a consequence of collecting a larger number of samples from multiple chains. We believe that the averaged prediction using many independent chains provides better generalization because many modes are likely to be explored (or, a large area of a single broad mode can be covered quickly if many chains reside there). After more investigation, we indeed observed that the same number of samples collected from a single chain (e.g. SGLD) cannot achieve the same level of accuracy obtained with multiple randomly initialized chains. Furthermore, we observed that given a lot more computational time, SGLD and DSGLD-S can approach the RMSE obtained by DSGLD-C as they also get a chance to explore other modes or to cover a larger area of a single mode. We will revisit the effect of multiple chains in more detail in the next section. Finally, note that Gibbs sampling achieves lower RMSE than DGSLD-C after around 20K seconds (5.5 hours) as shown in Fig. \ref{fig:netflix_d30} left (but the difference to DSGLD-C is very small). Note that for this dataset, $D$ and $L+M$ were not too large and we used 12-core single machine for parallel Gibbs sampling. Therefore the computational cost of each iteration was not extremely high.

\begin{figure*}[t]
\centering
	\subfigure[DSGLD-C on the Netflix dataset]
	{
		\hspace{-4.5mm}
		\includegraphics[width=0.35\textwidth]{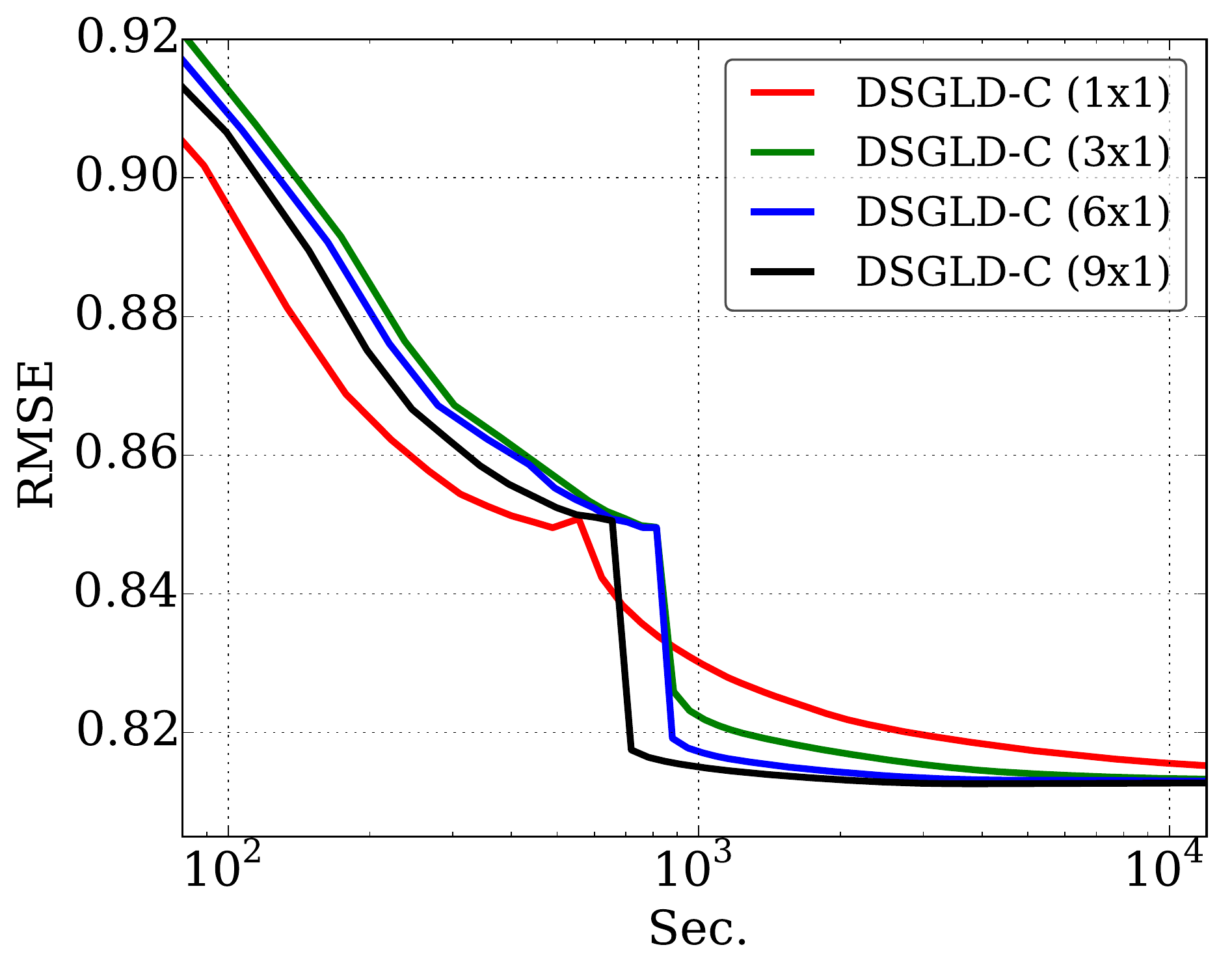}
	}
	\hspace{3.0cm}
	\subfigure[DSGLD-S on the Yahoo dataset]
	{
		\hspace{-4.5mm}
		\includegraphics[width=0.355\textwidth]{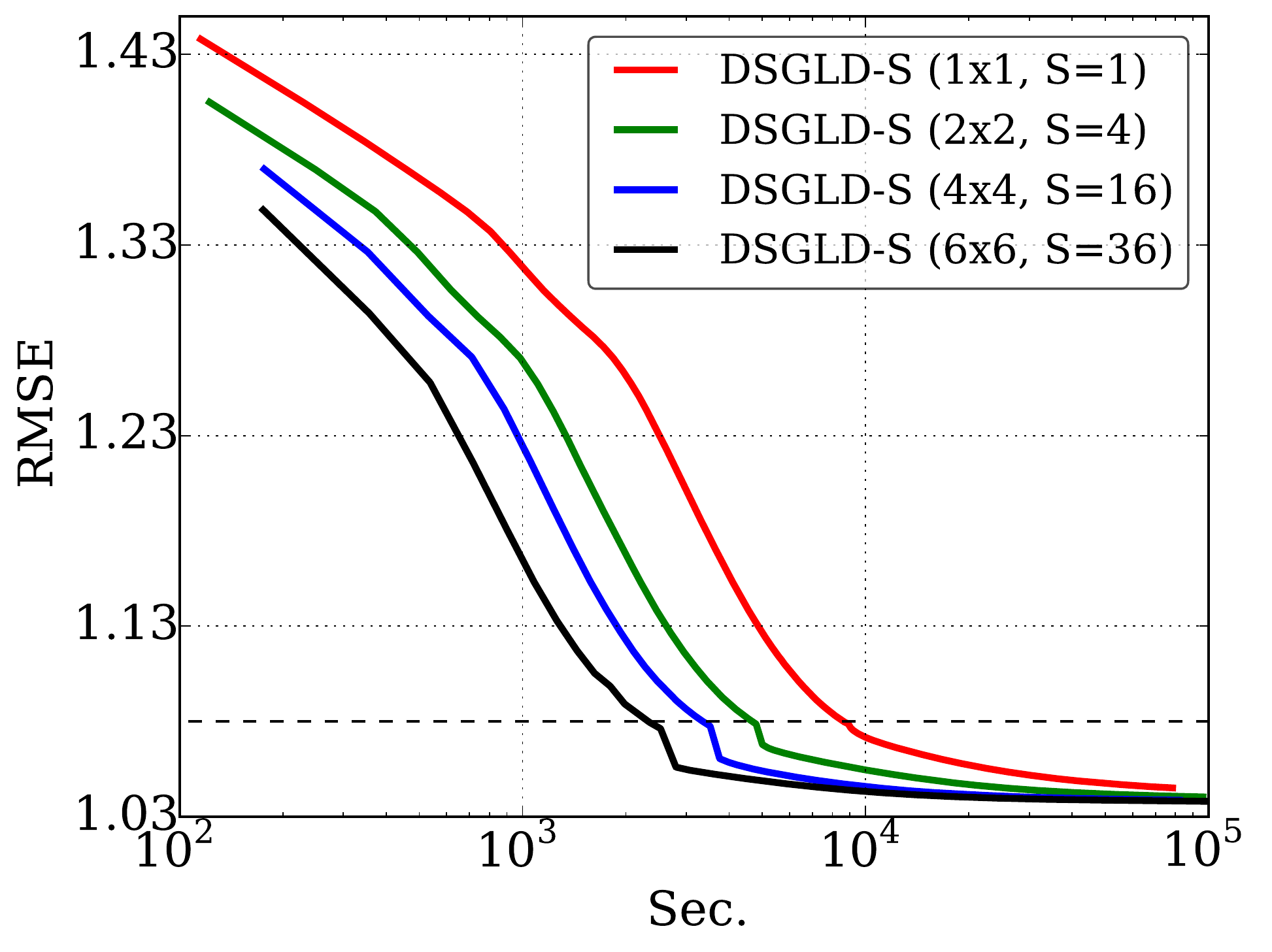}
	}
\caption{The effect of the number of chains, number of workers, and block split.} 
\label{fig:num_worker}
\end{figure*}

We present our results on the Yahoo dataset in Fig. \ref{fig:yahoo_d30} with $S=16$ workers. A remarkable point is that, here, unlike with the Netflix dataset, DSGLD-S outperforms DSGLD-C. This is because using orthogonal blocks increases the number of parameters updated per round, resulting in increased convergence speed even after offsetting the communication overhead.  As expected, a similar effect is observed for DSGD. The progress of parameter updates in DSGLD-C is relatively slow, requiring $S = 16$ rounds to update all the parameters. Besides, DSGLD-C has a much larger communication overhead because the full matrix $\bV$ has to be transferred between the parameter server and each of the workers, whereas only a small block of $\bV$ is transferred in DSGLD-S. Specifically, in DSGLD-C the parameter server sends and receives packets of total size $\cO((L+SM)D)$ per round whereas in DSGLD-S the total packet size is only $\cO((L+M)D)$. Although DSGLD-C is rather slow during burn-in, after burn-in we still see a faster decrease in RMSE compared to SGLD because multiple chains can mix better. Gibbs sampling converges slower than it does on the Netflix dataset because for the Yahoo dataset the number of latent vectors to update, i.e. $L+M$, increases by a factor of four, and the number of ratings, $N$, by a factor of seven.

For the Netflix dataset, after 1K seconds, DSGLD-C achieved the RMSE (0.8145) that the Gibbs sampler obtains at 10K seconds. Similarly, after 11K seconds, DSGLD-S achieved the RMSE (1.0454) that the Gibbs sampler obtains at 100K seconds. Therefore, the proposed method converges an order of magnitude faster than Gibbs sampling on both datasets, which is especially important when we only have a limited computational budget. 

DSGD converges to a prediction RMSE of 0.8462 on Netflix and 1.0576 on Yahoo after 1K seconds and 10K seconds respectively. Given the same amount of computational time, DSGLD achieves an error of 0.8161 on Netflix and 1.0465 on Yahoo, a relative improvement of 3.7\% and 1.1\%. After convergence, the RI increases to 4.1\% for Netflix and 1.8\% for Yahoo (See Table. \ref{tbl:D}).

% At the time when DSGD converged (around 1K seconds for Netflix and 10K seconds for Yahoo), the RMSE values of DSGLD were 0.8161 for Netflix and 1.0465 for Yahoo while DSGD converged to 0.8462 and 1.0576 respectively. Thus, we achieved a 3.7\% relative improvement for Netflix and an 1.1\% for Yahoo on the minimum computational budget. Given more time (up to 50K seconds for Netflix and 100K seconds for Yahoo), the relative improvements are increased to 4.1\% for Netflix and 1.8\% for Yahoo (See Table. \ref{tbl:D}).

\subsubsection{Number of chains and workers}
We also investigated the effect of the number of chains and the number of workers. The results are presented in Fig. \ref{fig:num_worker}. According to the observations from the previous experiment, we used DSGLD-C for Netflix and DSGD-S for Yahoo to study this effect. The latent feature dimension was set to $D$=30.

In Fig. \ref{fig:num_worker} (a), we compare DSGLD-C with $[1,3,6,9]$ chains (and workers) and in each case we evenly split the rows of the rating matrix between the chains. Note that DSGLD-C (1x1) is the same as SGLD running on a single-machine. We see that during burn-in DSGLD-C (1x1) converges faster than the other splits because there is no communication overhead. After burn-in, when the chains start averaging predictions, we see a sharp decrease in error for the other splits. Although splits with more chains decrease error much faster, they all eventually converge to a similar value. Due to poor mixing, a single chain (i.e. SGLD) converges very slowly.

In Fig. \ref{fig:num_worker} (b), we show results for DSGLD-S on the Yahoo dataset. We increased the number of workers to $[1,4,16, 36]$ to compare $[1,2,4,6]$ parallel chains.  Again DSGLD-S (1x1) denotes SGLD running on a single machine. We see that SGLD converges much more slowly because the dataset is larger than Netflix and SGLD has to update more parameters sequentially. Using more orthogonal blocks, DSGLD-S can update more parameters in parallel and we see more speed-up as we increase the number of workers. Although we increase the number of workers quadratically between the experiments, the packet size transferred between the parameter server and the workers stays constant at $\cO((L+M)D)$ because the block size also reduces accordingly. Even after burn-in (horizontal dotted black line at 1.08 RMSE) we see that with more chains we can decrease the error faster. This is because (i) multiple chains help to mix better by exploring a broader space (ii) each chain can mix faster by updating orthogonal blocks in parallel. 

\begin{figure*}[t]
\centering
	\subfigure[RMSE on Netflix]
	{
		\hspace{-2.5mm}
		\includegraphics[width=0.31\textwidth]{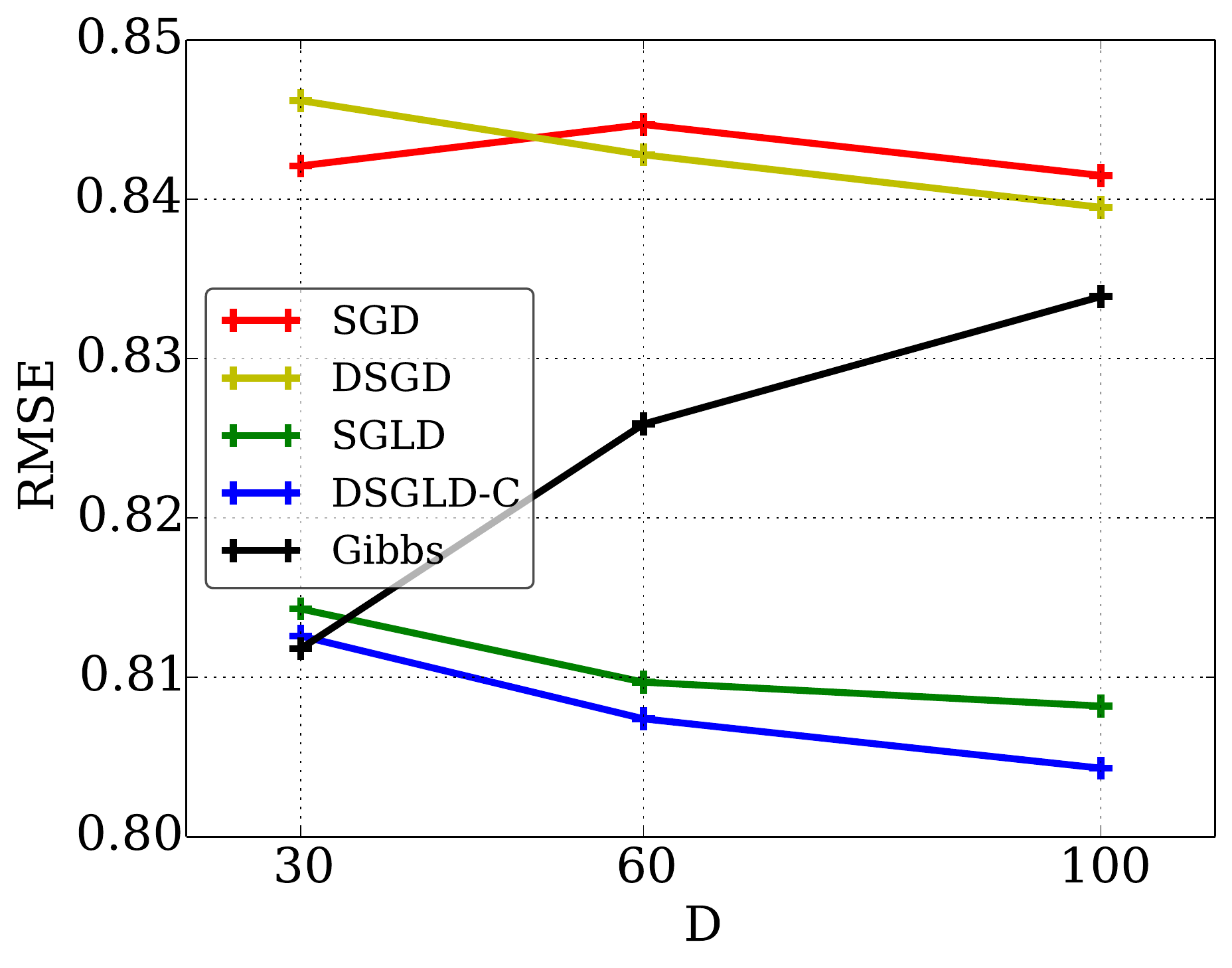}
	}
	\hspace{0.2cm}
	\subfigure[RMSE on Yahoo]
	{
		\hspace{-2.5mm}
		\includegraphics[width=0.317\textwidth]{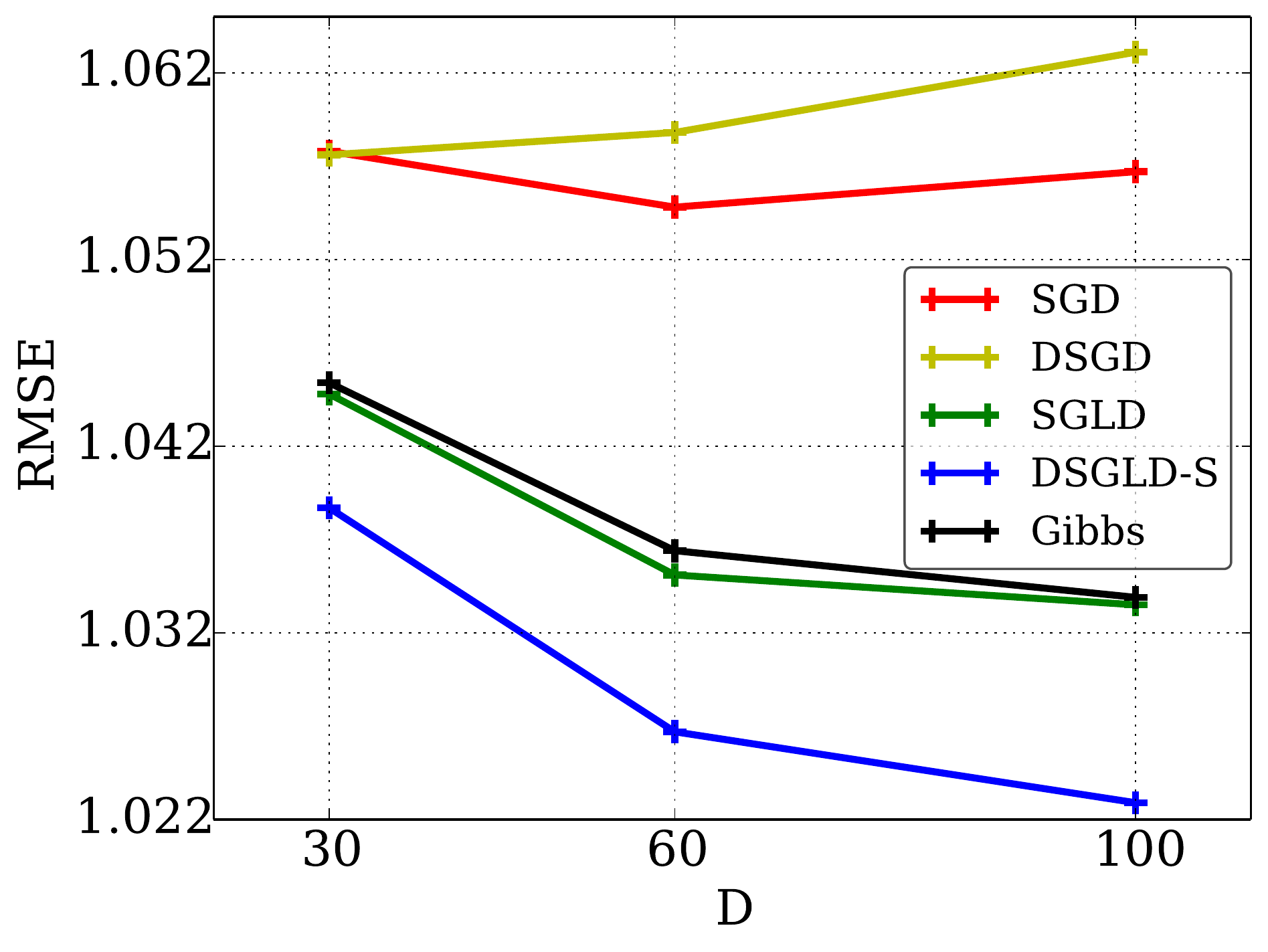}
	}
	\hspace{0.2cm}
	\subfigure[Required time per sample]
	{
		\hspace{-2.5mm}
		\includegraphics[width=0.31\textwidth]{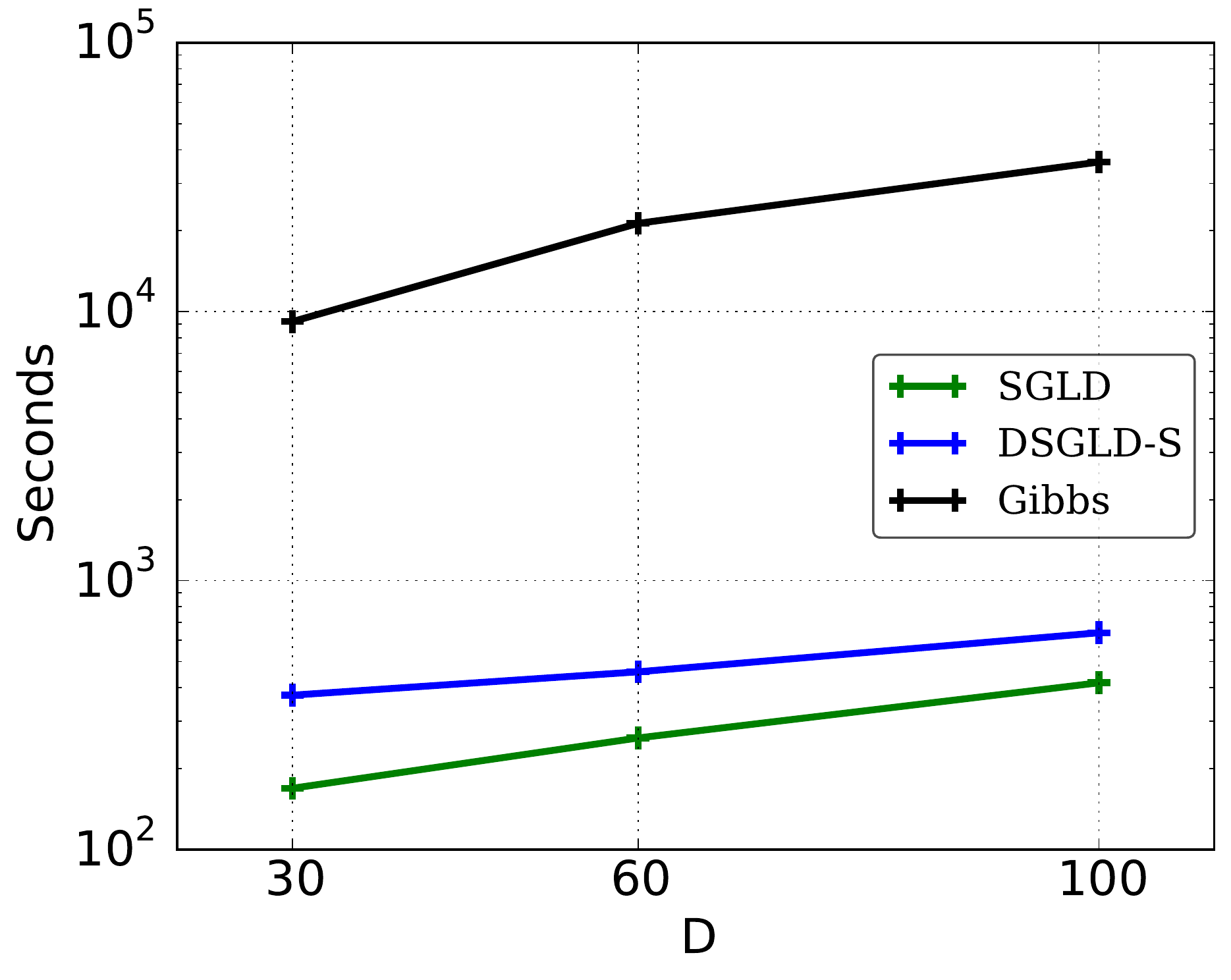}
	}
\vspace{-2.5mm}	
\caption{The effect of the latent feature dimension. (a) and (b) show RMSE for $D=[30,60,100]$ on (a) the Neflix dataset and (b) the Yahoo music ratings dataset. The maximum computational time was set to 50K seconds for Netflix and 100K ($D$=30), 200K ($D$=60), and 300K ($D$=100) seconds for Yahoo. (c) shows time (in seconds) required to draw a single sample on the Yahoo dataset.} 
\label{fig:D}
\end{figure*}

\begin{table*}[t]
	\centering
	\vspace{-0.0cm}
	\hspace{-0.0cm}
    	\begin{tabular}{ c | c  c  c  c  c }
     	D	& SGD     &  DSGD  & SGLD   & DSGLD-C & Gibbs\\ \hline\hline
    	30 	& 0.8421 & 0.8462  &  0.8143 &  0.8126 & \bf{0.8118}\\ 
    	        & -3.63\% & -4.13\%& -0.21\% & - & +0.09\%\\ \hline
	60    & 0.8447 & 0.8428	& 0.8097  & \bf{0.8074} & 0.8259\\ 
    	        & -4.62\% & -4.38\%& -0.28\% & - & -2.29\%\\ \hline	
    	100  & 0.8415 & 0.8395	& 0.8082  & \bf{0.8043} & 0.8339\\ 
    	        & -4.63\% & -4.37\%& -0.48\% & - & -3.68\%\\ 
    	\end{tabular}
	\hspace{1.0cm}
    	\begin{tabular}{ c | c  c  c  c  c }
     	D	& SGD     &  DSGD  & SGLD   & DSGLD-S  & Gibbs\\ \hline\hline
    	30 	& 1.0578 &  1.0576  & 1.0448 & \bf{1.0387} & 1.0454\\ 
    	        & -1.83\% & -1.82\%& -0.58 \%  & - & -0.64\%\\ \hline
    	60    & 1.0548 & 1.0588	& 1.0351  & \bf{1.0267} & 1.0364\\ 
    	        & -2.73\% & -3.13\%& -0.82\%  & - & -0.94\%\\ \hline
    	100  & 1.0567 & 1.0631	& 1.0335  & \bf{1.0229} & 1.0339\\ 
		& -3.30\% & -3.93\%& -1.04\%  & - & -1.08\%\\ 
    	\end{tabular}		
	\vspace{-0.0cm}
	\caption{RMSE and relative improvement (RI). Left: Netflix. Right: Yahoo. The percentage shown below each RMSE value is the relative improvement.}
	\label{tbl:D}	
	\vspace{-0.0cm}
\end{table*}

%%%%%%%%%%%%%%%%%%%%%%%%%%%%
\subsubsection{Latent feature dimension}

In Fig. \ref{fig:D}, we show how the latent feature dimension affects the final RMSE. The final RMSE on Netflix is measured after 50K seconds (14 hours) of computational time, because by then all algorithms had converged (except Gibbs sampling which is expected to take much longer). On the Yahoo dataset, we increased the computational time to 100K secs (1 day), 200K secs (2.3 days), and 300K secs (3.5 days) for $D$=[30,60,100], respectively, to give the Gibbs sampler more time to converge. In table \ref{tbl:D}, we show the RMSEs of the different algorithms and the \textit{relative} improvement (or deterioration) compared to DSGLD. The Relative Improvement (RI) of an algorithm $x$ is defined as $RI(x) = (r_d - r_x)/r_d$, where $r_x$ is the RMSE achieved by algorithm $x$ and $r_d$ is the RMSE obtained using DSGLD.

In both Fig. \ref{fig:D} (a) and (b), we see a large difference in performance between SG-MCMC (SGLD and DSGLD) and the optimization methods (SGD and DSGD). The RI is $3.6\% - 4.6\%$ on Netflix and $1.8\% - 3.9\%$ on the Yahoo dataset. As observed in \cite{ruslan08bpmf}, we see that the optimization methods do not consistently improve with increasing $D$. One reason is that optimization methods are highly sensitive to the hyperparameter values which become difficult to tune as the model becomes more complex. However, our method consistently improves as we increase $D$, because the hyper-parameters are sampled from their posterior distributions. We also see that the performance of Gibbs sampling on Netflix gets worse as $D$ increases, because we used the same amount of computational budget for all $D$ although the computation complexity increases as $D$ does. On the Yahoo dataset on which we increase computational time as we increase $D$, we see that the RMSE for Gibbs increases as $D$ increases, but is still lower than that of DSGLD.

In Fig. \ref{fig:D} (c), we compare the time (in seconds) required to draw a single sample for the three sampling algorithms at different values of $D$ on the Yahoo dataset. We see that Gibbs sampling is almost two orders of magnitude slower than SGLD. For $D$=100, SGLD, DSGLD-S, and Gibbs generated 688, 460, and 8 samples respectively in 300K seconds of computational time. For Netflix, Gibbs generated around 100 samples in 50K seconds  for $D$=30. Thus, even though the Gibbs sampler can produce higher quality samples (in terms of lower auto-correlation), the sampling speed is so slow that it cannot satisfactorily handle large scale datasets. 

%\section{Discussions}
%One may concern about the space complexity of the proposed algorithm because we need to store the samples. In fact, the proposed algorithm samples at much faster rate than Gibbs sampling does and thus requires to store larger number of samples. 
%
%We argue that this is not a problem in practice. First, in case that the testset is not so large that it can be predicted for a sample between the thinning interval, the parameter server, which can be also a cluster of many servers \cite{li14communication,Li13parameter}, can perform online-averaging of the predictions on the testset in background while performing the sampling. In this case, instead of storing many samples, we only need to manage a single set of averaged predictions on the testset. 
%
%Even when we need to store the samples however, it is not so difficult to manage. For example, in our experiment DSGLD converged almost fully with around 100 samples each of which (i.e. $\{\bU, \bV, \baa, {\bf b} \}$) is amount to about 1GB for $D$=60 in Yahoo dataset. Thus, the parameter server is required to store 100GB which is a trivial size to store in hard-disks as a background I/O. If the collected samples are stored in a distributed file system like MapReduce or Hadoop cluster, we can obtain the averaged prediction by a single map-reduce operation where each node is mapped to computing predictions on a subset of the samples which are then reduced at a master later to obtain the globally average.

\section{Conclusion}
Most applications of matrix factorization to recommender systems are based on stochastic gradient optimization algorithms because these are the only ones that can computationally handle very large datasets. However, by restricting ourselves to such simple algorithms, we miss out on all the advantages of Bayesian modelling such as quantifying uncertainty, controlling over-fitting, incorporating prior information and better prediction accuracy. In this paper, we introduced a novel algorithm for scalable distributed Bayesian matrix factorization that achieves the best of both worlds, i.e. it inherits all the advantages of Bayesian inference at the speed of stochastic gradient optimization.

Our algorithm, based on Distributed Stochastic Gradient Langevin Dynamics, uses only a mini-batch of ratings to make each update as in Stochastic Gradient Descent optimization. By running multiple chains in parallel, and also using multiple workers within a chain to update orthogonal blocks, we can scale up Bayesian Matrix Factorization to very large datasets. Parallel chains with different random initializations also help us to average predictions from multiple modes and improve accuracy. Moreover, our algorithm can effectively handle datasets that are distributed across multiple machines unlike traditional MCMC algorithms. 

We believe that our method is just one example of a much larger class of scalable distributed Bayesian matrix factorization methods. For example, we can consider using more sophisticated stochastic gradient algorithms \cite{AhnKorattikaraWelling12, PatTeh13sgrld, tianqi14sghmc, dingfang14sgnht} in place of SGLD to further improve the mixing rate. 

%ACKNOWLEDGMENTS are optional
\section*{Acknowledgments}
We thank Tianqi Chen and members of Yahoo labs personalization team for useful comments and discussions. This work is supported by NSF grant IIS-1216045 and Amazon AWS in Education Grant award. 

\bibliographystyle{abbrv}
\bibliography{Refs_sungjin}  % sigproc.bib is the name of the Bibliography in this case
% ACM needs 'a single self-contained file'!
%\balancecolumns
\appendix
%Appendix A
\section{Step-size Parameters}
\begin{table}[h]
	\centering
	\vspace{-0.0cm}
	\hspace{-0.0cm}
    	\begin{tabular}{ c | c  c  c  c  c }
     		     & SGD  &  DSGD & SGLD   & DSGLD-C & DSGLD-S \\ \hline\hline
    	$\ep_0$& 9e-6  & 1e-6     &  9e-6    & 9e-6           & 3e-6 \\ 
	$\ka$    & 50     & 10        &   1000   & 1000          & 500\\ \hline
%	round length & & & 50 & & 
    	\end{tabular}
	\vspace{-0.0cm}
	\caption{Stepsize parameters for Netflix $D$=30 and 9 workers}
	\label{tbl:}	

	\vspace{1.0cm}

	\centering
	\hspace{-0.0cm}
    	\begin{tabular}{ c | c  c  c  c  c }
     		     & SGD    &  DSGD  & SGLD   & DSGLD-C  & DSGLD-S\\ \hline\hline
    	$\ep_0$& 1.5e-6  & 3e-7     &  1.5e-6  & 9e-7          & 1.5e-6 \\ 
	$\ka$    & 500      & 100       &   1000   & 1000         & 500\\ 
    	\end{tabular}
	\vspace{-0.0cm}
	\caption{Stepsize parameters for Yahoo $D$=30 and 16 workers}
	\label{tbl:}	
	\vspace{-0.0cm}
\end{table}

\end{document}